%% file: main.tex
\documentclass[runningheads]{llncs}
\usepackage{graphicx}
\usepackage{comment}
\usepackage{amsmath,amssymb} % define this before the line numbering.
\usepackage{color, colortbl}
\usepackage{booktabs}
\usepackage{xcolor}
\usepackage{multirow}
\usepackage{comment}

\usepackage{xspace}
\usepackage{wrapfig}
\usepackage{sidecap}
\usepackage{microtype}

\usepackage[width=122mm,left=12mm,paperwidth=146mm,height=193mm,top=12mm,paperheight=217mm]{geometry}

\newcommand*{\belowrulesepcolor}[1]{%
  \noalign{%
    \kern-\belowrulesep
    \begingroup
      \color{#1}%
      \hrule height\belowrulesep
    \endgroup
  }%
}
\newcommand*{\aboverulesepcolor}[1]{%
  \noalign{%
    \begingroup
      \color{#1}%
      \hrule height\aboverulesep
    \endgroup
    \kern-\aboverulesep
  }%
}

\definecolor{lightgray}{gray}{0.95}

\newcommand{\our}{$\boldsymbol{\star}$\xspace}
\renewcommand{\paragraph}[1]{{\vspace{3pt} \noindent \bf #1}}

\def\cf{\emph{cf.}\xspace}
\def\ie{\emph{i.e.}\xspace}
\def\eg{\emph{e.g.}\xspace}

\def\etal{\emph{et al.}\xspace}
\def\etc{\emph{etc.}\xspace}

\def\roxf{$\mathcal{R}$Oxford\xspace}
\def\rox{$\mathcal{R}$Oxf\xspace}

\def\rpar{$\mathcal{R}$Paris\xspace}
\def\rpa{$\mathcal{R}$Par\xspace}

\def\rmil{$\mathcal{R}$1M\xspace}

\def\ours{LAttQE\xspace}
\def\oursdba{LAttDBA\xspace}

\def\mq{\mathbf{q}}
\def\md{\mathbf{d}}
\def\mtq{\mathbf{\tilde{q}}}
\def\mhq{\mathbf{\hat{q}}}
\newcommand{\mdi}[1]{\mathbf{d}_{#1}}
\newcommand{\mtdi}[1]{\mathbf{\tilde{d}}_{#1}}

\mathchardef\mhyphen="2D
\mathchardef\mplus=\mathcode`+

\input{plots}

\usepackage[pagebackref=true,breaklinks=true,colorlinks,bookmarks=false]{hyperref}

\begin{document}

% equation spacing
\setlength{\abovedisplayskip}{3pt}
\setlength{\belowdisplayskip}{3pt}

% \linenumbers
\pagestyle{headings}
\mainmatter
\def\ECCVSubNumber{6054}  % Insert your submission number here

\title{Attention-Based Query Expansion Learning} % Replace with your title

\begin{comment}
\titlerunning{ECCV-20 submission ID \ECCVSubNumber} 
\authorrunning{ECCV-20 submission ID \ECCVSubNumber} 
\author{Anonymous ECCV submission}
\institute{Paper ID \ECCVSubNumber}
\end{comment}
%******************

%\begin{comment}
\titlerunning{Attention-Based Query Expansion Learning}

\author{Albert Gordo \and Filip Radenovic \and Tamara Berg}
% \author{Albert Gordo\orcidID{0000-0001-9229-4269} \and Filip Radenovic\orcidID{0000-0002-7122-2765} \and Tamara Berg\orcidID{0000-0002-1272-3359}}
%
\authorrunning{Gordo \etal}
\institute{Facebook AI}
%\end{comment}
%******************
\maketitle

\begin{abstract}
Query expansion is a technique widely used in image search consisting in combining highly ranked images from an original query into an expanded query that is then reissued, generally leading to increased recall and precision.
An important aspect of query expansion is choosing an appropriate way to combine the images into a new query.
Interestingly, despite the undeniable empirical success of query expansion, ad-hoc methods with different caveats have dominated the landscape, and not a lot of research has been done on \emph{learning} how to do query expansion.
In this paper we propose a more principled framework to query expansion, where one trains, \emph{in a discriminative manner}, a model that learns how images should be aggregated to form the expanded query. Within this framework, we propose a model that leverages a self-attention mechanism to effectively learn how to transfer information between the different images before aggregating them.
Our approach obtains higher accuracy than existing approaches on standard benchmarks. More importantly, our approach is the only one that consistently shows high accuracy under different regimes, overcoming caveats of existing methods.

\end{abstract}

%\vspace{5pt}
\section{Introduction}
\label{sec:intro}

Image search is a fundamental task in computer vision, directly applied in a number of applications such as visual place localization~\cite{kalantidis2011viral,sattler2012image,arandjelovic2016netvlad}, 3D reconstruction~\cite{heinly2015reconstructing,schonberger2016structure,makantasis2016wild}, content-based image browsing~\cite{weyand2011discovering,mikulik2013image,alletto2016exploring}, \etc
Image search is typically cast as a nearest neighbor search problem in the image representation space, originally using local feature matching and bag-of-words-like representations~\cite{sivic2003video}, and, more recently, CNN-based global image representations~\cite{gordo2017learning,alphaqe}.

To increase the accuracy of image search systems, a robust representation of the query image is desirable. Query expansion~(QE) is a commonly used technique to achieve this goal, where relevant candidates produced during an initial ranking are aggregated into an expanded query, which is then used to search more images in the  database. 
Aggregating the candidates reinforces the information shared between them and injects new information not available in the original query.
This idea was originally exploited in the work of Chum~\etal~\cite{chum2007total}, introducing the first attempt at image retrieval QE.
This averaging of query and top ranked results~\cite{chum2007total}, or ad-hoc variations of it~\cite{chum2011total2,tolias2014visual,dqe,gordo2017learning,alphaqe}, are now used as a standard method of performance boosting in image retrieval.

Selecting which images from the initial ranking should be used in the QE procedure is however a challenging problem, since we do not have guarantees that they are actually relevant to the query.
Early methods use strong geometrical verification of local features to select true positives~\cite{chum2007total,chum2011total2,dqe,tolias2014visual}. 
As CNN-based global features lack this possibility, the most common approach is to use the $k$-nearest neighbors to the query~\cite{gordo2017learning,alphaqe},  potentially including false positives.
Yet, if $k$ is larger than the number of relevant images, topic drift will degrade the results significantly. This leads to two unsatisfying alternatives: either use a very small $k$, potentially not leveraging relevant images, or use weighted average approaches with decreasing weights as a function of ranking~\cite{gordo2017learning} or image similarity~\cite{alphaqe}, where setting the appropriate decay is a task  just as challenging as choosing the optimal $k$.  This has unfortunately led to many works tuning the $k$ parameter directly on test, as well as to use different values of $k$ for each dataset.
Replacing $k$-nearest neighborhoods with similarity-based neighborhoods turn out to be just as unstable, as, unlike inlier count for local features, cosine similarity of CNN global features is not directly comparable between different query images~\cite{noh2017large}.

\begin{figure}[t!]
    % \vspace{-10pt}
    \centering
    \includegraphics[width=0.85\textwidth]{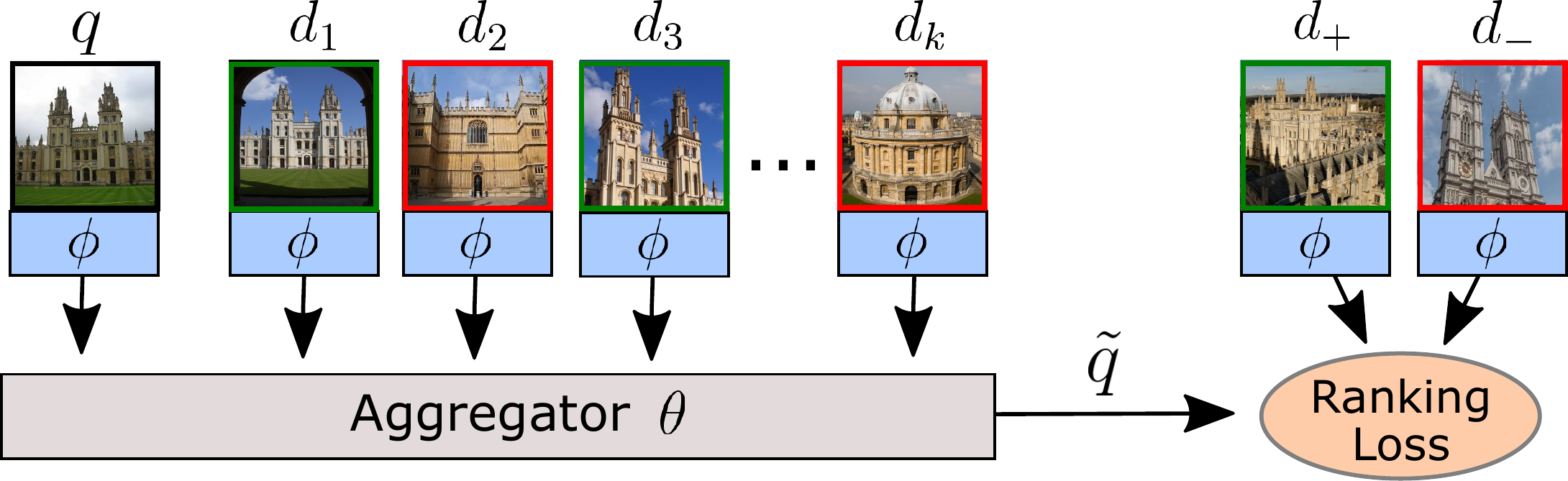}
    % \vspace{-10pt}
    \caption{Outline of our proposed approach. During training, we sample a query $q$ and its nearest neighbors in the training dataset 
    (where their features have been precomputed with the function $\phi$, typically a CNN) 
    and use our proposed attention-based model $\theta$ to aggregate them into an expanded query $\tilde{\mq}$. Given positive ($\md_\mplus$) and/or negative ($\md_\mhyphen$) samples, we use a ranking loss to optimize $\theta$. Images with the {\color{greennn}\textbf{green}} ({\color{reddd}\textbf{red}}) border represent relevant (non-relevant) samples to the query. At inference, we construct the expanded $\tilde{\mq}$ given ${\mq}$ and its neighbors in the index, and use it to query the index again.
    \label{fig:fig1}
    % \vspace{-10pt}
    }
\end{figure}

We argue that existing QE approaches are generally not robust and use ad-hoc aggregation methods, and instead propose to cast QE as a discriminative learning problem.  
Similar to recent methods that learn embeddings suitable for image retrieval using large-scale datasets \cite{gordo2017learning,alphaqe}, we formulate the problem as a ranking one, where we train an aggregator that produces the expanded query, optimized to rank relevant samples ahead of non-relevant ones, \cf Figure \ref{fig:fig1}.
We use a large-scale dataset, disjoint from the evaluation ones, to train and validate our model and its parameters.
We then leverage a self-attention mechanism to design an aggregator model that can transfer information between the candidates (Figure \ref{fig:fig2}), enabling the model to learn the importance of each sample before aggregating them. We call this model Learnable Attention-based Query Expansion, or \ours.
Unlike previous QE approaches, \ours does not produce monotonically decreasing weights, allowing it to better leverage the candidates in the expansion. 
\ours is more robust to the choice of $k$ thanks to the large-scale training, which  enables the model to better handle false positive  amongst the top neighbors, and is usable across a wide range of class distributions without sacrificing the performance at any number of relevant images.

Our contributions are as follows: 
(i) We show that standard query expansion methods, albeit seemingly different, can be cast under the same mathematical framework, allowing one to compare their advantages and shortcomings in a principled way. 
(ii) We propose to treat query expansion as a discriminative learning problem, where an aggregation model is learned in a supervised manner.
(iii) We propose \ours, an aggregation model designed to share information between the query and the top ranked items by means of self-attention. 
We extend this query expansion model to also be useful for database-side augmentation.
(iv) We show that our proposed approach outperforms commonly-used query expansion methods in terms of both accuracy and robustness on standard benchmarks.

\section{Related work}
\label{sec:rw}

\paragraph{Image retrieval query expansion.}
Average query expansion (AQE) in image retrieval was originally proposed for representations based on local features~\cite{chum2007total}, and tuned for the bag-of-words search model~\cite{sivic2003video}, where local features are aggregated after a strict filtering step, usually based on strong feature geometry~\cite{chum2007total,chum2011total2} or Hamming embedding distance~\cite{tolias2014visual}.
For CNN-based global image representation, AQE is implemented by mean-aggregating the top $k$ retrieved images~\cite{gordo2017learning,alphaqe}.
It has been argued that setting an optimal $k$ for several datasets of different positive image distributions is a non-trivial task~\cite{alphaqe}.
Instead, Gordo~\etal~\cite{gordo2017learning} propose using a weighted average, where the weight is a monotonically decaying function over the rank of retrieved images. We denote this method as average query expansion with decay, or AQEwD.
Likewise, Radenovic~\etal~\cite{alphaqe} use a weighted average, where the weights  are computed as a power-normalized similarity between the query and the top ranked images.
This method, known as alpha query expansion ($\alpha$QE), has proven to be fairly robust to the number of neighbors $k$, and is used as a \textit{de facto} standard by a number of recent state-of-the-art image retrieval works~\cite{radenovic2018revisiting,revaud2019listwise,gu2019gmp,husain2019actnet,fan2019retrieval,husain2019remap}.
Finally, Arandjelovic~\etal~\cite{dqe} proposed discriminative query expansion (DQE) where they train a linear SVM using top ranked images as positives, and low ranking images as negatives, and use the resulting classifier as the expanded query. 
Note that this is very different from our method, as DQE trains independent classifiers for each query, while we train one single model using a large disjoint dataset.

\paragraph{Image retrieval database pre-processing.}
If the database is fixed at indexing time, one can pre-process the database to refine the image representations and improve the accuracy accuracy.
Database-side augmentation~(DBA)~\cite{dqe} is a method that applies QE to each image of the database and replaces the original representation of the image by its expanded version.
Although it increases the offline pre-processing time, it does not increase the memory requirements of the pipeline or the online search time.
All aggregation-based QE methods described in the previous paragraph~\cite{chum2007total,gordo2017learning,dqe,alphaqe} can be applied as different flavors of DBA, including our proposed \ours.
A different line of work~\cite{qin2011hello,shen2013spatially,delvinioti2014image} indexes local neighborhoods of database images together with their respective representations, in order to refine the search results based on the reciprocal neighborhood relations between the query and database images.
Besides offline pre-processing, these approaches require additional storage and are slower at query time.
Finally, some works~\cite{iscen17diffusion,chang2019explore} build a nearest neighbor graph using the database image representations and traverse it at query time, or, alternatively, encode graph information into image descriptors~\cite{liu2019guided}.
It increases the amount of required memory by storing the graph structure of the database, and increases online search complexity by orders of magnitude.
Both reciprocal-nearest-neighbor and graph-based methods are complementary to our work, and can be applied after augmenting the database representations with our method.
When dealing with dynamically-growing indexes, applying these methods becomes even more challenging, which makes them generally unappealing despite the accuracy~gains.

\paragraph{Self-attention.} 
The self-attention transformer \cite{vaswani2017attention} has established itself as the core component of strong language representation models such as BERT \cite{devlin2019bert} or GPT-2 \cite{radford2019gpt} due to its ability to capture complex interactions between tokens and due to how easy it is to increase the capacity of models simply by stacking more encoders.
Self-attention has also shown applications outside of NLP. Wang \etal \cite{wang2018nonlocal} leverage self-attention to aggregate descriptors from different parts of the image in order to capture interactions between them in a non-local manner. In a similar way, Girdhar and Ramanan \cite{girdhar2017} use self-attention as an approximation for second order pooling. In a different context, Lee \etal \cite{lee2019gp} use self-attention as a graph pooling mechanism to combine both node features and graph topology in the pooling. 
In this paper we use self-attention as a way to transfer information between the top $k$ results so we can construct a more discriminative query. As we describe in Section~\ref{sec:qel}, self-attention is an excellent mechanism to~this~end.

\paragraph{Query expansion and relevance feedback in information retrieval.}
The information retrieval community has leveraged query expansion techniques for several decades \cite{maron60qe,rocchio1971rf,azad2019queryexpansion}. Most interestingly, in the information retrieval community, query expansion methods expand or reformulate query terms independently of the query and results returned from it, via, \eg, reformulation with a thesaurus \cite{manning2008intro}. What the image search community denotes as query expansion is generally known as relevance feedback (RF), and more precisely, pseudo-RF, as one generally does not have access to the true relevance of the neighbors -- although a case could be made for geometrical verification methods \cite{chum2007total} providing explicit feedback. Our focus in this work is not on information retrieval methods for two reasons: (i) they generally deal with explicit or implicit RF instead of pseudo-RF; (ii) they generally assume high-dimensional, sparse features (\eg bags of terms), and learn some form of term weighting that is not applicable in our case.

\vspace{-3pt}
\section{Attention-based query expansion learning}
\label{sec:qel}
\vspace{-3pt}

We start this section by presenting a generalized form of query expansion, and by showing that well-known query expansion methods can be cast under this framework. We then propose a general framework for learning query expansion in a discriminative manner. Last, we propose \ours (Learnable Attention-Based Query Expansion), an aggregation model that leverages self attention to construct the augmented query and that can be trained within this framework.

\vspace{-5pt}
\subsection{Generalized query expansion}

We assume that there exists a known function $\phi: \Omega \rightarrow \mathcal{R}^D$ that can embed items (\eg images) into an $l_2$-normalized $D$-dimensional vectorial space. For example, $\phi$ could be a CNN trained to perform image embedding \cite{gordo2017learning,alphaqe,revaud2019listwise}.
Let us denote with $q$ a query item, and, following standard convention of using bold typeface for vectors, let us denote with $\mq = \phi(q)$ its $D$-dimensional embedding. Similarily, let us denote with $\{\md\}^k = \mdi{1}$, $\mdi{2}$, \ldots, $\mdi{k}$ the embeddings of the top $k$ nearest neighbors of $\mq$ in a dataset $\mathcal{D}$ according to some measure of similarity, \eg the cosine similarity, and sorted in decreasing order.
Let us also denote with $\{\md\}^\mhyphen$ a collection of dataset items that are not close to the query, according to the same measure of similarity.
Last, for convenience, let us alias $\mdi{0}:=\mq$.

We propose the following generalized form of query expansion: 
\begin{equation}
    \mhq = \frac{1}{Z} \sum_{i=0}^k \theta(\mdi{i} \: |\:\mq, \:\{\md\}^k, \:\{\md\}^\mhyphen, \:i),
\label{eq:generalized}
\end{equation}
where $Z$ is a normalization factor, and $\theta$ is a learnable function that takes an individual sample and applies a transformation conditioned on the original query $\mq$, the top $k$ retrieved results $\{\md\}^k$ , a collection of low-ranked samples $\{\md\}^\mhyphen$, and its position $i$ in the ranking. The final augmented query is computed by aggregating the transformed top $k$ results, including the query, and applying a normalization $Z$ (\eg $\ell_2$ normalization)\footnote{Note that Eqn. \eqref{eq:generalized} does not aggregate over $\{\md\}^\mhyphen$. This is just to ease the exposition; negative samples can also be aggregated if the specific method requires it, \eg, DQE.}.

Standard query expansion methods can be cast under this framework. In fact, they can be cast under a more constrained form: $\theta(\mdi{i} \: |\:\mq, \:\{\md\}^k, \:\{\md\}^\mhyphen, \:i) = w_i \mdi{i}$, where the value of $w_i$ is method-dependent, see Table~\ref{tab:methods}.
Two things are worth noticing. 
First, for all methods, $w_i$ depends either on positional information (\eg the sample got ranked at position $i$ out of $k$, as done by AQEwD), or on information about the content (\eg the power-normalized similarity between the item and the query, as done by $\alpha$QE). None of the methods leverage both the positional and the content information simultaneously. 
Second, except for DQE, all methods produce a monotonically decreasing $\mathbf{w}$, \ie, if $i > j$, then $w_i \leq w_j$. 
The implication is that these methods do not have the capacity to uplift the samples amongst the top $k$ retrieved results that are indeed relevant to the query but were ranked after some non-relevant samples. That is, any top-ranked, non-relevant item will contribute more to the construction of the expanded query than any relevant item ranked after it, with clear negative consequences.

\begin{table}[t!]
\resizebox{0.96\textwidth}{!}{\begin{minipage}{\textwidth}
\footnotesize
\centering
\begin{tabular}{lll}
\toprule
\multicolumn{2}{l}{\textbf{Method}} & $\theta(\mdi{i} \: |\:\mq, \:\{\md\}^k, \:\{\md\}^\mhyphen, \:i) = w_i \mdi{i}$\\
\midrule
 \cite{chum2007total} & \textbf{AQE:} Average QE & $w_i = 1$\\
\specialrule{.01em}{0.1em}{0.1em}
\cite{gordo2017learning} & \textbf{AQEwD:} AQE with decay \;\;& $w_i = (k - i)/k$\\
\specialrule{.01em}{0.1em}{0.1em}
\cite{dqe} & \textbf{DQE:} Discriminative QE & $\mathbf{w}$ is the dual-form solution of an SVM optimization\\
 & &  problem using $\{\md\}^k$ as positives and $\:\{\md\}^\mhyphen$ as negatives\\
\specialrule{.01em}{0.1em}{0.1em}
 \cite{alphaqe} & \textbf{$\boldsymbol{\alpha}$QE:} $\alpha$-weighted QE & $w_i = \text{sim}(\mq, \mdi{i})^\alpha$,  \\
 && with $\alpha$ being a hyperparameter. \\
\bottomrule
\end{tabular}
\end{minipage}}
\vspace{5pt}
\caption{Standard query expansion (QE) methods and their associated transformations. More details about the methods can be found in Section~\ref{sec:rw}.
\label{tab:methods}
\vspace{-10pt}
}
\end{table}

\subsection{Query expansion learning}

We propose that, following recent approaches in representation learning \cite{alphaqe,gordo2017learning}, one can learn a differentiable $\theta$ transformation in a data-driven way (Figure \ref{fig:fig1}). This training is done in a supervised manner, and ensures that relevant items to the (expanded) query are closer to it than elements that are not relevant. This is achieved by means of losses such as the triplet loss \cite{weinberger2009metric} or the contrastive loss \cite{hadsell2006contrastive}.
The approach requires access to an annotated dataset (\eg rSfM120k \cite{alphaqe}), but the training data and classes used to learn $\theta$ can be disjoint from the pool of index images that will be used during deployment, as long as the distributions are similar. 
From that point of view, the requirements are similar to other existing image embedding learning methods in the literature.

At training time, besides sampling queries, positive, and negative samples, one also has to consider the nearest neighbors of the query for the expansion. Sampling a different subset of neighbors each time, as a form of data augmentation, can be useful to improve the model robustness. We provide more details about the process in the experimental section.
Finally, we note that this framework allows one to learn $\theta$ and $\phi$ jointly, as well as to learn how to perform QE and DBA jointly, but we consider those variations out of the scope of this work.

\subsection{Learnable Attention-based Query Expansion (\ours)}

We propose a more principled $\theta$ function that overcomes the caveats of previous methods, and that can be trained using the framework described in the previous section.
In particular, our $\theta$ function is designed to be capable of transferring information between the different retrieved items, giving all top-ranked relevant samples the opportunity to significantly contribute to the construction of the expanded query.
To achieve this we rely on a self-attention mechanism. We leverage the transformer-encoder module developed by Vaswani \etal \cite{vaswani2017attention}, where, in a nutshell, a collection of inputs first share information through a multi-head attention mechanism, and later are reprojected into an embedding space using fully connected layers with layer normalization and residual connections -- see Fig.~1 of Vaswani \etal \cite{vaswani2017attention} for a diagram of this module (left) and the decoder module (right), not used in this work.
Stacking several of these encoders increases the capacity of the model and enables sharing more contextual information.
The exact mechanism that the stack of self-attention encoders uses to transfer information is particularly suited for our problem: 

 1. The encoder's scaled dot-product attention~\cite{vaswani2017attention} performs a weighted sum  of the form $\sum_{j=0}^k \text{Softmax}(\mdi{i}^T\left[ \mdi{0},\mdi{1},\ldots,\mdi{k}\right]/C)_j \mdi{j}$, where $C$ is a constant,
in practice computing the similarity between $\mdi{i}$ and all other inputs and using that as weights to aggregate all the inputs. Observing equations \eqref{eq:generalized} and \eqref{eq:qel}, one can see self-attention as a way to perform expansion of the input samples, leading to richer representations that are then used to compute the weights.

 2. The multihead attention enables focusing on different parts of the representations. This is important because computing similarities using only the original embedding will make it difficult to  change the original ranking.
By using multihead attention, we discover parts of the embeddings that are still similar between relevant items and dissimilar between non-relevant items, permitting the model to further upweight relevant items and downweight  non-relevant ones.

 3. Under this interpretation of the encoder, the stack of encoders allows the model to ``refine'' the expansion process in an iterative manner. One can see this as expanding the queries, making a first search, using the new neighbors to expand a better query, find new neighbors, etc. Although the pool of neighbors remains constant, we expect the expansion to become more and more accurate.

\paragraph{Aggregation.} 
The stack of encoders takes the  query $\mq$ and the top results $\mdi{1}\ldots \mdi{k}$ as input, and produces outputs $\mtq$ and $\mtdi{1}\ldots \mtdi{k}$.
To construct the expanded query, a direct solution consists in aggregating them (\eg through average or weighted average) into a single vector that represents the expanded query. 
However, this is challenging in practice, as it requires the encoder to learn how to create outputs that lie in the same space as the original data, something particularly hard when the embedding function $\phi$ is not  being simultaneously learned. We empirically verify that learning such a function leads to weak results. Although we speculate that learning a ``direct'' $\theta$ function jointly with $\phi$ could lead to superior results, the practical difficulties involved in doing so make this approach unappealing.
Instead, to ensure that we stay in a similar space, we relax the problem and also construct the expanded query as a weighted sum of the top $k$ results, where the weights $\mathbf{w}$ are predicted by our model. If we denote with $M$ the stack of encoders, the transformed outputs can be represented as 
\begin{equation}
    \mtdi{i} = M(\{\mq\} \cup \{\md\}^k)_i.
\end{equation}
Then, inspired by other methods such as $\alpha$QE, we can construct the weight $w_i$ as the similarity between item $\mdi{i}$ and the query $\mq$ \emph{in the transformed space}, \ie, $w_i = \text{sim}(\mtq, \mtdi{i})$. This leads to our proposed $\theta$:

\begin{equation}
    \theta(\mdi{i} \: |\:\mq, \:\{\md\}^k, \:\{\md\}^\mhyphen, \:i) = \text{sim}(\mtq, \mtdi{i}) \mdi{i}.
\label{eq:qel}    
\end{equation}

\begin{SCfigure}[][t!]
    \centering
    \includegraphics[width=0.48\linewidth]{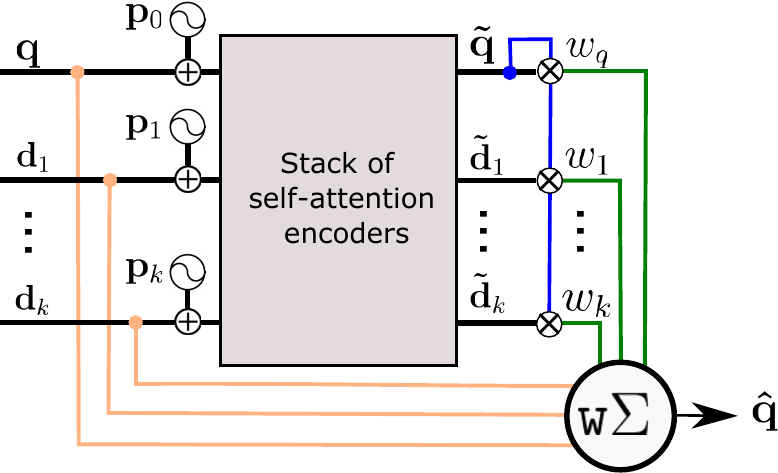}
    \caption{Proposed aggregator. The output $\mathbf{\hat{q}}$ is constructed as the weighted sum ($\texttt{w}\Sigma$) of the query $\mq$ and the nearest neighbors $\mdi{1} \ldots \mdi{k}$. The weights are computed by running the inputs through a stack of self-attention encoders after including positional information (\protect\includegraphics[height=0.25cm]{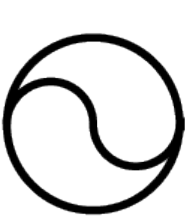}) and computing the similarity (through a normalized dot product $\otimes$) between the transformed query $\mtq$ and all the transformed samples $\mtdi{1}\ldots\mtdi{k}$.
    \label{fig:fig2}
    }
\end{SCfigure}

\paragraph{Including rank information.} As presented, the proposed method does not leverage in any way the ranking of the results. Indeed, the encoders see the inputs as a set, and not as a sequence of results. This prevents the model from leveraging this information, \eg by learning useful biases such as ``top results tend to be correct, so pay more attention to them to learn the transformations''. To enable the model to reason not only about the content of the results but also about their ranking, we follow standard practice when dealing with transformers and include a positional encoding that is added to the inputs before being consumed by the encoder, \ie, $\text{pe}(\mdi{i}) = \mdi{i} + \mathbf{p}_i$, where each $\mathbf{p}_i \in \mathcal{R}^D$ is a learnable variable within our model. 
The full proposed aggregator that leverages $\theta$ with positional encoding is depicted in Figure \ref{fig:fig2}.

\paragraph{Auxiliary classification loss.} Since, at training time, we have access to the annotations of the images, we know which of the top $k$ results are relevant to the query and which ones are not.
This enables us to have an auxiliary linear classifier that predicts whether $\mtdi{i}$ is relevant to the query or not. 
The role of this classifier, which is only used at train time and discarded at inference time, is to encourage the relevant and non-relevant outputs of the encoder to be linearly separable, inducing the relevant items to be more similar to the query than the non-relevant ones. Our empirical evaluation in Section \ref{sec:experiments} shows that the use of this auxiliary loss can noticeably increase the accuracy of the model. 

\vspace{-5pt}
\subsection{Database-side augmentation}
\vspace{-3pt}

Database-side augmentation (DBA) is technique complementary to query expansion. Although different variations have been proposed \cite{turcot2009dba,dqe,tolias2015selective,gordo2017learning}, the main idea is that one can perform query expansion, \emph{offline}, on the database images. 
This produces an expanded version of the database images, which are then indexed, instead of indexing the original ones. 
When issuing a new query, one searches on the expanded index, and not on the original~one. 

Our proposed approach can also be used to perform better database-side augmentation, using $\theta$ to aggregate the top $k$ neighbors of each database image.
However, this approach did not work in practice. We believe that the reason is that, on the database side, many images are actually distractors, unrelated to any query, and our model was assigning weights too high for unrelated images when using them as queries. To address this, we propose to use a tempered softmax over the weights, \ie, instead of computing our weights as $w_i = \text{sim}(\mtq, \mtdi{i})$, we compute it as
\begin{equation}
w_i = \text{Softmax}(\text{sim}(\mtq, [\mtdi{0},\mtdi{1},\ldots,\mtdi{k}] ) / T)_i,
\end{equation}
where $\text{sim}(\mtq, [\mtdi{0},\mtdi{1},\ldots,\mtdi{k}])$  is a vector of similarities between $\mtq$ and all the $\mathbf{\tilde{d}}$s, and $T$ is a learnable scalar. 
 
To achieve the best results, we employ a curriculum learning strategy, where first we train our model without softmax, and then we freeze the parameters of the model, incorporate the tempered softmax, and continue training while updating only $T$. This strategy led to a DBA that not only gave the best results in terms of accuracy but that was also more stable than other variants.

\section{Experiments}
\label{sec:experiments}

In this section we discuss implementation details of our training, evaluate different components of our method, and compare to the state of the art. 

\subsection{Training setup and implementation details}

\paragraph{Image representation.} 
For all experiments we use a publicly-available, state-of-the-art model for image retrieval~\cite{alphaqe}\footnote{\href{https://github.com/filipradenovic/cnnimageretrieval-pytorch}{github.com/filipradenovic/cnnimageretrieval-pytorch}} to extract the underlying features.
We use the best-performing model from the project page (trained on Google Landmarks 2018 data~\cite{noh2017large}), consisting of a ResNet101 trunk followed by generalized-mean pooling and a whitening layer, which produces features of 2048 dimensions.
Following~\cite{alphaqe}, we extract features at 3 scales ($1, \sqrt{2}, 1/\sqrt{2}$), mean-aggregate them, and finally $\ell_2$-normalize to form the final 2048D representation.

\paragraph{Training dataset.} 
We use the publicly available rSfM120k created by Radenovic~\etal~\cite{alphaqe}, which comprises images selected from 3D reconstructions of landmarks and urban scenes.
These reconstructions are obtained from an unordered image collection using a combined local-feature-based image retrieval and structure-from-motion pipeline.
The 3D reconstruction cluster ids serve as a supervision for selecting positive and negative pairs.
In total, 91642 images from 551 classes are used for training, while additional 6403 database images -- 1691 of which are used as queries -- from 162 classes, disjoint from the training ones, are set aside for validation. 
Performance on validation is measured as mean average precision (mAP)~\cite{philbin2007object} over all 1691 queries.

\paragraph{Learning configuration.} To train \ours we follow \cite{alphaqe} and use a contrastive loss of the form $yz^2+(1-y)max(0,m-z)^2$, with m being the margin, $z=||\hat{q}-d||$, and $y\in \{0, 1\}$ denotes whether d is relevant to q or not. We backpropagate through $\hat{q}$, which in turn optimizes the transformers (see Fig \ref{fig:fig2}). Other recent ranking losses \cite{deng2019arcface,revaud2019listwise,ng2020solar} could also be used. 
Since the base representations are already strong, we use a margin of $0.1$, which ensures that positives are pulled together while only pushing away negatives that are too close to the query. \ours consists of a stack of 3 transformer encoders, each one with $64$ heads. We did not see any improvement after further increasing the capacity of the model. The self-attention and fully-connected layers within the encoders preserve the original dimensionality of the inputs, 2048D.
We also follow \cite{alphaqe} regarding the sampling strategy for positives and negatives: we select 5 negatives per positive, found in a pool of 20000 samples that gets refreshed every 2000 updates. 
When sampling neighbors to construct the augmented query, as a form of data augmentation, the exact number of neighbors is drawn randomly between 32 and 64, and neighbors are also randomly dropped according to a Bernoulli distribution (where the probability of dropping neighbors in each query is itself drawn from a uniform distribution between 0 and 0.6).
The auxiliary  classification head uses a binary cross-entropy loss.
We use Adam to optimize the model, with a batch size of 64 samples, a weight decay of $1e\mhyphen6$, and an initial learning rate of $1e\mhyphen4$ with an exponential decay of $0.99$. The optimal number of epochs (typically between 50 and 100) is decided based on the accuracy on the validation set, and is typically within $1\%$ of the optimal iteration if it was validated directly on test. 

\vspace{5pt}
\subsection{Test datasets and evaluation protocol}

\paragraph{Revisited Oxford and Paris.}
Popular Oxford Buildings~\cite{philbin2007object} and Paris~\cite{philbin2008lost} datasets have been revisited by Radenovic~\etal~\cite{radenovic2018revisiting}, correcting and improving the annotation, adding new more difficult queries, and updating the evaluation protocol. 
Revisited Oxford (\roxf) and Revisited Paris (\rpar) datasets contain 4,993 and 6,322 images respectively, with 70 held out images with regions of interest that are used as queries.
Unlike the original datasets, where the full-size version of query images are present in the database side, this is not the case in revisited versions, making query expansion a more challenging task. 
For each query, the relevant database images were labeled according to the ``difficulty'' of the match. 
The labels are then used to define three evaluation protocols for \roxf and \rpar: Easy (E), Medium (M), and Hard (H).
As suggested by Radenovic~\etal~\cite{radenovic2018revisiting}, which points out that the Easy protocol is saturated, we only report results on the Medium and Hard protocols.
Note that Oxford and Paris landmarks are not present in rSfM120k training and validation datasets.

\paragraph{Distractors.}
A set of 1 million hard distractor images (\rmil) were collected in~\cite{radenovic2018revisiting}.
These distractors can, optionally, be added to both \roxf and \rpar to evaluate performance on a more realistic large-scale setup.

\paragraph{}We do not evaluate on INRIA Holidays~\cite{jegou2008hamming}, another common retrieval dataset, since performing query expansion on Holidays is not a standard practice.

%\vspace{5pt}
\subsection{Model study}

Table \ref{tab:study} displays the results of our proposed model, using all components (row~ii), and compares it with the results without query expansion (row~i). We use 64 neighbors for query expansion, as validated on the validation set of rSfM120k.
Our model clearly improves results on \roxf and \rpar, both on the M and H settings.  
We further study the impact of the components introduced in Sec.~\ref{sec:qel}.

\begin{table}[ht!]
\centering
\scriptsize
\def\cw{0.925cm}
\newcolumntype{L}[1]{>{\raggedright\let\newline\\\arraybackslash\hspace{0pt}}m{#1}}
\newcolumntype{C}[1]{>{\centering\let\newline\\\arraybackslash\hspace{0pt}}m{#1}}
\newcolumntype{R}[1]{>{\raggedleft\let\newline\\\arraybackslash\hspace{0pt}}m{#1}}
\def\arraystretch{1.0} % 1 is the default, row height
\begin{tabular}{L{0.6cm}L{4.8cm}C{\cw}C{\cw}C{\cw}C{\cw}C{\cw}}
\toprule
 && \multicolumn{2}{c}{\roxf} & \multicolumn{2}{c}{\rpar}  \\
 \cmidrule(lr){3-4} \cmidrule(lr){5-6}
  && M & H & M & H & Mean\\
  \midrule
  (i) & No QE & 67.3 & 44.3 & 80.6 & 61.5& 63.4\\
  \midrule
  (ii) & \textbf{Full model} & 73.4 & 49.6 & 86.3 & 70.6 & 70.0\\
  \midrule
  (iii) & Without self-attention & 66.0 & 41.5 & 86.1 &70.2 &  66.0\\
  (iv) & Without positional encoding & 58.6 & 33.2 & 87.8 & 73.4 & 63.2 \\
  (v) & Without visual embedding &  67.1 & 42.9 & 83.8 &66.7 & 65.1 \\
  (vi) & Without auxiliary loss & 71.8 & 47.0 & 85.8 & 69.4 & 68.5\\
  \bottomrule
\end{tabular}
\vspace{5pt}
\caption{Mean average precision (mAP) performance of the proposed model (ii) compared to the baseline without query expansion (i) and to variations where parts of the model have been removed (iii-vi).}
\label{tab:study}
\end{table}

\paragraph{Self-attention:} replacing the stack of self-attention encoders  with a stack of fully-connected layers leads to a very noticeable drop in accuracy (iii), highlighting how important the attention is for this model.

\paragraph{Positional encoding (PE):} Removing the PE (iv) leads to a very pronounced loss in accuracy for \roxf (which has very few relevant images per query).
PE is necessary for queries with few relevant items because the model has to learn which images are important, and anchoring to the query (through the PE) enables it to do so. This is less important for queries with many relevant items, as in \rpar.
We additionally experiment with a position-only setup (v), where the self-attention computes the weights using only the positional encodings, not the actual image embeddings. This leads to a content-unaware weighting function, such as the AQE or AQEwD methods. The drop in accuracy is also remarkable, highlighting the need to combine both content and positional information.

\paragraph{Auxiliary loss:} Removing the auxiliary loss (vi) leads to a small but consistent drop in accuracy. 
Although the model is fully functional without this auxiliary loss, it helps the optimization process to find better representations.

\paragraph{Inference time:} When considering 64 neighbors for the expansion, our non-optimized PyTorch implementation can encode, on average, about 250 queries per second on a single Tesla M40 GPU. 
This does not include the time to extract the query embedding, which is orders of magnitude slower than our method (about 4 images per second on the same GPU) and the main bottleneck.
Techniques such as  distillation \cite{sanh2019distilbert} and quantization \cite{shen2020qbert}, that have worked for transformer-based models, could further increase speed and reduce memory~use.

\subsection{Comparison with existing methods}

\paragraph{Query expansion (QE).}
We compare the performance of our proposed method with existing QE approaches.
All methods and their associated transformations are given in Table~\ref{tab:methods}.
For \ours, hyper-parameters are tuned on the validation set of rSfM120k, that has no overlapping landmarks or images with the test datasets.
For competing methods, we select their hyper-parameters on the mean performance over test datasets, giving them an advantage.
We denote the number of neighbors used for QE as nQE.
\textbf{AQE:} nQE=2; 
\textbf{AQEwD:} nQE=4; \linebreak
\textbf{$\boldsymbol{\alpha}$QE:} nQE=72, $\alpha$=3; 
\textbf{DQE:} nQE=4, neg=5, $C=0.1$;
\textbf{\ours:} nQE=64.

\paragraph{Database-side augmentation (DBA).} 
All of the before-mentioned methods can be combined with DBA. We separately tune all hyper-parameters in this combined scenario.
We denote number of neighbors used for DBA as nDBA.
\textbf{ADBA+AQE:} nDBA=4, nQE=4;
\textbf{ADBAwD+AQEwD:} nDBA=4, nQE=6;
\textbf{$\boldsymbol{\alpha}$DBA+$\boldsymbol{\alpha}$QE:} nDBA=36, nQE=10, $\alpha$=3;
\textbf{DDBA+DQE:} nDBA=4, nQE=2, $C$=0.1, neg=5;
\textbf{\oursdba{+}\ours:} nDBA=48, nQE=64.

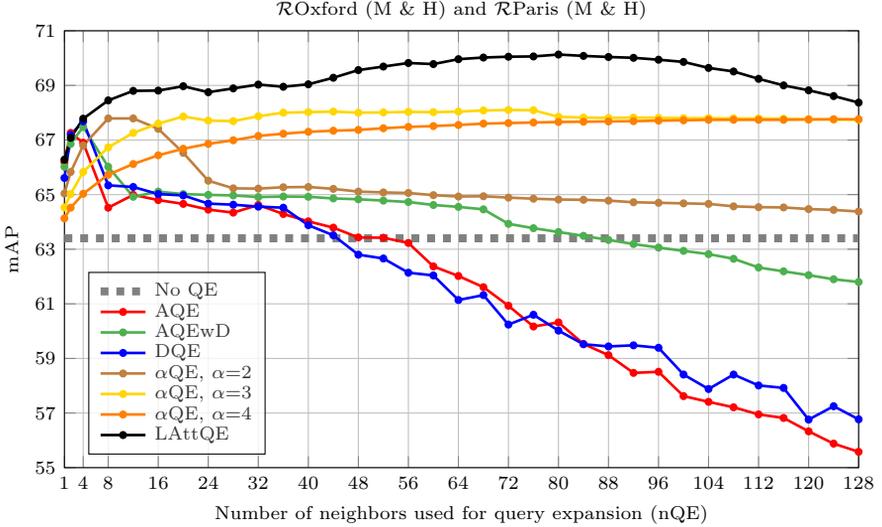
\begin{figure}[t]
\input{fig_map_nqe_trends}
% \vspace{-10pt}
\caption{Mean average precision over all queries of four protocols (\roxf (M \& H) and \rpar (M \& H)) as a function of the number of neighbors used for query~expansion. 
\label{fig:map_nqe_trends}
}
\end{figure}

\paragraph{Sensitivity to the number of neighbors used in the QE.}
Figure \ref{fig:map_nqe_trends} shows the mean accuracy of \ours as well as other query expansion methods on \roxf and \rpar, as a function of the number of neighbors used in the expansion. 
We highlight: 
(i) Unsurprisingly, methods that assume all samples are positive (\eg AQE, DQE) degrade very fast when the number of neighbors is not trivially small. AQEwD degrades a bit more gracefully, but can still obtain very bad results if nQE is not chosen carefully.  
(ii) It is also unsurprising that $\alpha$QE has become a standard, since the accuracy is high and results do not degrade when nQE is high. However, this only happens because of the weighting function is of the form $r^\alpha$, with $r < 1$, \ie, the weight rapidly converges to zero, and therefore most neighbors barely have any impact in the aggregation.
(iii) Our proposed \ours consistently obtains the best results across the whole range of nQE. Our method is not limited by a weight that converges to zero, and therefore can still improve when $\alpha$QE has essentially converged (nQE $>40$).

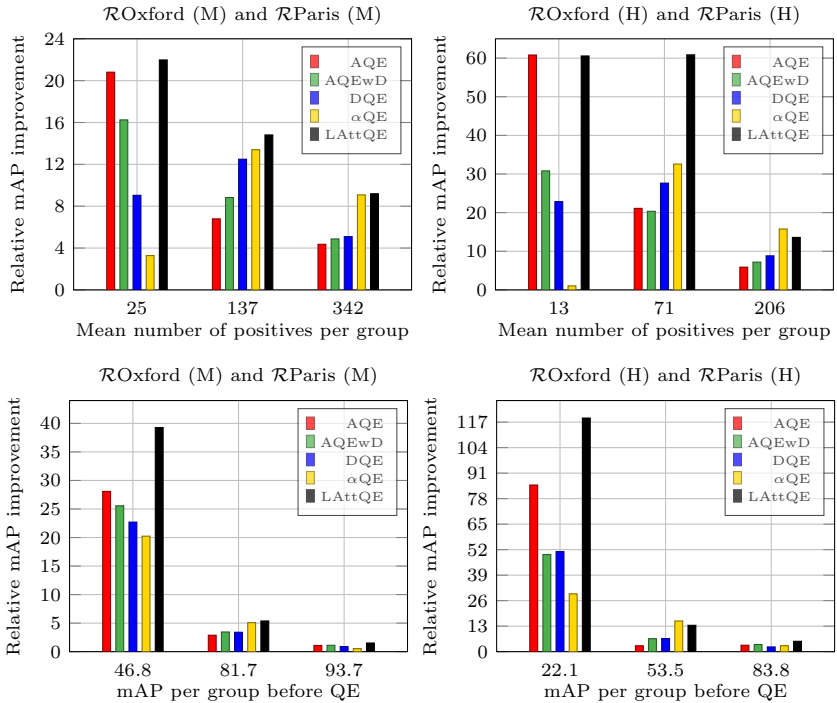
\begin{figure}[t]
\pgfplotsset{compat=1.11,
    /pgfplots/ybar legend/.style={
    /pgfplots/legend image code/.code={%
       \draw[##1,/tikz/.cd,yshift=-0.15em]
        (0cm,0cm) rectangle (3pt,0.5em);},
   },
}
\input{fig_map_np_split3}
\input{fig_map_ap_split3}
% \vspace{-10pt}
\caption{Relative mean average precision (mAP) improvement at different number of relevant images (top) and AP regimes (bottom) split into 3 groups. Evaluation performed on \roxf and \rpar at two difficulty setups, Medium (left) and Hard (right). Mean number of relevant images over all queries in the group (top) and mean average precision over all queries in the group (bottom) shown under respective group's bar plot. 
\label{fig:map_np_split3}
}
\end{figure}

\paragraph{Different ``number of relevant images'' and ``AP'' regimes.} 
We evaluate query expansion impact at different regimes to showcase further differences between methods. 
In all cases we report the relative improvement in mAP introduced by using query expansion. 
In the first set of experiments, see Figure~\ref{fig:map_np_split3} (top), we group queries based on the number of relevant images, using percentiles 33 and 66 as cut-off. 
AQE (with nQE=4) works very well for queries with very few relevant samples, but leads to small improvements when the number of relevants is high, as they are not leveraged. 
On the other hand, $\alpha$QE, with $\alpha$=3 and nQE=72 obtains good results when the number of relevants is high, but really struggles when the number of relevants is low. 
\ours is the only method that is able to obtain high accuracy on all regimes. 
Figure~\ref{fig:map_np_split3} (bottom) groups queries based on their accuracy before query expansion. Similarly, \ours is the only method that consistently obtains high accuracy. 
%Also, as expected, the relative gains for queries with high accuracy / high number of relevant images are generally smaller than those with low accuracy / low number of relevant images.

\paragraph{State-of-the-art comparison.}
Table \ref{tab:baseline_comparison} reports the accuracy of different methods on \roxf and \rpar, both with and without the \rmil distractor set. The optimal number of neighbors for our approach (64 for \ours and 48 for \oursdba) was decided on the validation set of rSfM120k. 
On the other hand, the optimal number of neighbors for the remaining methods was adjusted on test to maximize their mean accuracy on \roxf and \rpar, giving them an unfair edge. 
Our method is the only one that consistently obtains good results on both \roxf and \rpar. 
Compare this to other methods, where, for example, $\alpha$QE obtains the best results on \rpar but the worst results on \roxf, while AQE obtains the best results on \roxf (excepting our method) but the worst results on \rpar.
Generally, this gap becomes even larger when including the \rmil distractors. When using DBA and QE we observe the same trends: although some method can be slightly more accurate on specific datasets, our approach is the only one that obtains consistently good results on all datasets.

\begin{table}[t]
\centering
\scriptsize
\def\cw{0.83cm}
\newcolumntype{L}[1]{>{\raggedright\let\newline\\\arraybackslash\hspace{0pt}}m{#1}}
\newcolumntype{C}[1]{>{\centering\let\newline\\\arraybackslash\hspace{0pt}}m{#1}}
\newcolumntype{R}[1]{>{\raggedleft\let\newline\\\arraybackslash\hspace{0pt}}m{#1}}
\begin{tabular}{L{0.45cm}L{3cm}C{\cw}C{\cw}C{\cw}C{\cw}C{\cw}C{\cw}C{\cw}C{\cw}C{\cw}}
\toprule
 & & \multicolumn{2}{c}{\rox} & \multicolumn{2}{c}{\rox~+~\rmil} & \multicolumn{2}{c}{\rpa} & \multicolumn{2}{c}{\rpa~+~\rmil} & \\
\cmidrule(lr){3-4} \cmidrule(lr){5-6} \cmidrule(lr){7-8} \cmidrule(lr){9-10}
  & & M & H & M & H & M & H & M & H & Mean \\
\midrule
\belowrulesepcolor{lightgray}
\rowcolor{lightgray} \multicolumn{2}{l}{\textbf{No QE}}  &  &  &  &  &  &  &  &  & \\
\aboverulesepcolor{lightgray}
 \midrule
\multicolumn{2}{l}{---}  & 67.3 & 44.3 & 49.5 & 25.7 & 80.6 & 61.5 & 57.3 & 29.8 &  52.0\\
\midrule
\belowrulesepcolor{lightgray}
\rowcolor[gray]{0.95} \multicolumn{2}{l}{\textbf{QE}}  &  &  &  &  &  &  &  &  & \\
\aboverulesepcolor{lightgray}
\midrule
\cite{chum2007total} & AQE & 72.3 & 49.0 & 57.3 & 30.5 & 82.7 & 65.1 & 62.3 & 36.5 & 56.9 \\
\cite{gordo2017learning} & AQEwD & 72.0 & 48.7 & 56.9 & 30.0 & 83.3 & 65.9 & 63.0 & 37.1 & 57.1 \\
\cite{dqe} & DQE & 72.7 &  48.8 & 54.5 & 26.3 & 83.7 & 66.5 & 64.2 & 38.0 &  56.8 \\
\cite{alphaqe} & $\alpha$QE & 69.3 & 44.5 & 52.5 & 26.1  & \textbf{86.9} & \textbf{71.7}  & 66.5 & 41.6 &  57.4 \\
\midrule
\our & \ours & \textbf{73.4} &  \textbf{49.6} & \textbf{58.3} & \textbf{31.0} & 86.3 & 70.6 & \textbf{67.3} & \textbf{42.4} &  \textbf{59.8} \\
\midrule
\belowrulesepcolor{lightgray}
\rowcolor[gray]{0.95} \multicolumn{2}{l}{\textbf{DBA + QE}}  & & & & & & & & & \\
\aboverulesepcolor{lightgray}
\midrule
\cite{chum2007total} & ADBA + AQE & 71.9 & 53.6 & 55.3 & 32.8 & 83.9 & 68.0 & 65.0 & 39.6 & 58.8 \\
\cite{gordo2017learning} & ADBAwD + AQEwD & 73.2 & 53.2 & 57.9 & 34.0 & 84.3 & 68.7 & 65.6 & 40.8 & 59.7 \\
\cite{dqe} & DDBA + DQE & 72.0 & 50.7 & 56.9 & 32.9 & 83.2 & 66.7 & 65.4 & 39.1 &  58.4 \\
\cite{alphaqe} & $\alpha$DBA + $\alpha$QE & 71.7 & 50.7 & 56.0 & 31.5 & 87.5 & 73.5 & \textbf{70.6} & \textbf{48.5} &  61.3 \\
\midrule
\our & \oursdba + \ours  & \textbf{74.0} & \textbf{54.1} & \textbf{60.0} & \textbf{36.3} & \textbf{87.8} & \textbf{74.1} & 70.5 & 48.3  & \textbf{63.1}\\
\bottomrule
\end{tabular}
\vspace{10pt}
\caption{
Performance evaluation via mean average precision (mAP) on \roxf (\rox) and \rpar (\rpa) with and without 1 million distractors (\rmil). Our method is validated on validation part of rSfM120k and is marked with \our. Other methods are validated directly on mAP over all queries of 4 protocols of \roxf and \rpar. 
\label{tab:baseline_comparison}
}
\end{table}

\section{Conclusions}
\label{sec:conclusions}
In this paper we have presented a novel framework to learn how to perform query expansion and database side augmentation for image retrieval tasks. Within this framework we have proposed \ours, an attention-based model  that outperforms commonly used query expansion techniques on standard benchmark while being more robust on different regimes.
Beyond \ours, we believe that the main idea of our method, tackling the aggregation for query expansion as a supervised task learned in a discriminative manner, is general and novel, and hope that more methods build on top of this idea, proposing new aggregation models that lead to more efficient and accurate search systems.

\newpage
\clearpage
\bibliographystyle{splncs04}
\bibliography{egbib}

\end{document}

%% file: plots.tex
\usepackage{pgfplots}
\usepackage{pgfplotstable}
\usepackage{pgf,tikz}
\pgfplotsset{compat=1.14}

\newcommand{\extdata}[1]{\input{#1}}
\newcommand{\leg}[1]{\addlegendentry{#1}}

\definecolor{greenn}{rgb}{0.30,0.69,0.31}
\definecolor{greennn}{rgb}{0.10,0.50,0.10}
\definecolor{yelloww}{rgb}{1.0000,0.8392,0}
\definecolor{reddd}{rgb}{0.70,0.20,0.20}

%% file: fig_map_nqe_trends.tex
\centering
\input{data}
\begin{tikzpicture}
\begin{axis}[%
	ylabel near ticks, yticklabel pos=left,
	xlabel near ticks,
	font=\scriptsize,
	width=1.0\linewidth,
	height=0.6\linewidth,
	xlabel={Number of neighbors used for query expansion (nQE)},
	ylabel={mAP},
	xlabel style  = {yshift = 0pt},
	title={\roxf~(M \& H) and \rpar~(M \& H)},
	legend pos=south west,
    legend style={cells={anchor=west}, font =\scriptsize, fill opacity=0.8, row sep=-2.5pt},	
    ymax = 71,
    ymin = 55,
    xmin = 1,
    xmax = 128,
    grid=both,
    xtick={1,4,8,16,24,...,128},
    ytick={1, 3, ..., 100},
    title style = {yshift = -5pt}
]  
    \addplot[color=gray, dashed, mark=None, mark size=1, line width=3] table[x=nqe, y expr={63.4}] \nqetrends; \leg{No QE};
    \addplot[color=red, solid, mark=*, mark size=1, line width=1.0] table[x=nqe, y expr={\thisrow{avg}}] \nqetrends; \leg{AQE};
    \addplot[color=greenn, solid, mark=*, mark size=1, line width=1.0] table[x=nqe, y expr={\thisrow{avgpdecay}}] \nqetrends; \leg{AQEwD};
    \addplot[color=blue, solid, mark=*, mark size=1, line width=1.0] table[x=nqe, y expr={\thisrow{dqe}}] \nqetrends; \leg{DQE};
    \addplot[color=brown, solid, mark=*, mark size=1, line width=1.0] table[x=nqe, y expr={\thisrow{aqe2}}] \nqetrends; \leg{$\alpha$QE, $\alpha$=2};
    \addplot[color=yelloww, solid, mark=*, mark size=1, line width=1.0] table[x=nqe, y expr={\thisrow{aqe3}}] \nqetrends; \leg{$\alpha$QE, $\alpha$=3};
    \addplot[color=orange, solid, mark=*, mark size=1, line width=1.0] table[x=nqe, y expr={\thisrow{aqe4}}] \nqetrends; \leg{$\alpha$QE, $\alpha$=4};
    \addplot[color=black, solid, mark=*, mark size=1, line width=1.0] table[x=nqe, y expr={\thisrow{qel}}] \nqetrends; \leg{\ours};
\end{axis}
\end{tikzpicture}

%% file: fig_map_np_split3.tex
\pgfplotsset{
  bar cycle list/.style={
    cycle list={%
      {red!50!black,fill=red,mark=none},%
      {greenn!50!black,fill=greenn,mark=none},%
      {blue!50!black,fill=blue,mark=none},%
      {yelloww!50!black,fill=yelloww,mark=none},%
      {black!50!black,fill=black,mark=none},%
    }
  },
}
\centering
\begin{tabular}{cc}
\begin{tikzpicture}
\begin{axis}[
    ybar,
    font=\scriptsize,
    title = {\roxf(M) and \rpar(M)},
    title style={yshift=-1ex},          
    width=0.5\linewidth,
    height=0.4\linewidth,
    legend style={opacity = 0.7, font=\tiny, row sep = -2pt, mark options={scale=3.5}, xshift = 0pt, cells={align=right}},
    legend cell align={right},
    legend pos=north east,
    grid=both,
    ylabel={Relative mAP improvement},
    xlabel={Mean number of positives per group},
    symbolic x coords={A, B, C},
    xtick=data,
    xticklabels={$25$, $137$, $342$},
    % nodes near coords,
    % nodes near coords align={vertical},
    /pgf/bar width=3pt,% bar width
    xtick style={draw=none},
    ymin = 0, ymax = 24,
    ytick={0,4,...,100},
    x tick label style={inner sep=0pt},   
    enlarge x limits=0.3,
    ]
\addplot coordinates {(A, 20.81) (B, 6.79) (C, 4.36)};
\addplot coordinates {(A, 16.25) (B, 8.83) (C, 4.87)};
\addplot coordinates {(A, 9.04) (B, 12.49) (C, 5.09)};
\addplot coordinates {(A, 3.28) (B, 13.40) (C, 9.08)};
\addplot coordinates {(A, 21.98) (B, 14.81) (C, 9.19)};
\legend{AQE, AQEwD, DQE, $\alpha$QE, \ours}
\end{axis}
\end{tikzpicture}
& 
\begin{tikzpicture}
\begin{axis}[
    ybar,
    font=\scriptsize,
    title = {\roxf(H) and \rpar(H)},
    title style={yshift=-1ex},          
    width=0.5\linewidth,
    height=0.4\linewidth,
    legend style={opacity = 0.7, font=\tiny, row sep = -2pt, mark options={scale=3.5}, xshift = 0pt, cells={align=right}},
    legend cell align={right},
    legend pos=north east,
    grid=both,
    ylabel={Relative mAP improvement},
    xlabel={Mean number of positives per group},
    symbolic x coords={A, B, C},
    xtick=data,
    xticklabels={$13$, $71$, $206$},
    % nodes near coords,
    % nodes near coords align={vertical},
    /pgf/bar width=3pt,% bar width
    xtick style={draw=none},
    ymin = 0, ymax = 65,
    ytick={0,10,...,100},
    x tick label style={inner sep=0pt},   
    enlarge x limits=0.3,
    ]
\addplot coordinates {(A, 60.80) (B, 21.11) (C, 5.89)};
\addplot coordinates {(A, 30.81) (B, 20.35) (C, 7.19)};
\addplot coordinates {(A, 22.85) (B, 27.64) (C, 8.82)};
\addplot coordinates {(A, 1.05) (B, 32.58) (C, 15.77)};
\addplot coordinates {(A, 60.56) (B, 60.85) (C, 13.59)};
\legend{AQE, AQEwD, DQE, $\alpha$QE, \ours}
\end{axis}
\end{tikzpicture}
\end{tabular}

%% file: fig_map_ap_split3.tex
\pgfplotsset{
  bar cycle list/.style={
    cycle list={%
      {red!50!black,fill=red,mark=none},%
      {greenn!50!black,fill=greenn,mark=none},%
      {blue!50!black,fill=blue,mark=none},%
      {yelloww!50!black,fill=yelloww,mark=none},%
      {black!50!black,fill=black,mark=none},%
    }
  },
}
\centering
\begin{tabular}{cc}
\begin{tikzpicture}
\begin{axis}[
    ybar,
    font=\scriptsize,
    title = {\roxf(M) and \rpar(M)},
    title style={yshift=-1ex},          
    width=0.5\linewidth,
    height=0.4\linewidth,
    legend style={opacity = 0.7, font=\tiny, row sep = -2pt, mark options={scale=3.5}, xshift = 0pt, cells={align=right}},
    legend cell align={right},
    legend pos=north east,
    grid=both,
    ylabel={Relative mAP improvement},
    xlabel={mAP per group before QE},
    symbolic x coords={A, B, C},
    xtick=data,
    xticklabels={$46.8$, $81.7$, $93.7$},
    % nodes near coords,
    % nodes near coords align={vertical},
    /pgf/bar width=3pt,% bar width
    xtick style={draw=none},
    ymin = 0, ymax = 44,
    ytick={0,5,...,100},
    x tick label style={inner sep=0pt},   
    enlarge x limits=0.3,
    ]
\addplot coordinates {(A, 28.08) (B, 2.88) (C, 1.09)};
\addplot coordinates {(A, 25.56) (B, 3.45) (C, 1.12)};
\addplot coordinates {(A, 22.70) (B, 3.40) (C, 0.90)};
\addplot coordinates {(A, 20.24) (B, 5.08) (C, 0.54)};
\addplot coordinates {(A, 39.27) (B, 5.38) (C, 1.52)};
\legend{AQE, AQEwD, DQE, $\alpha$QE, \ours}
\end{axis}
\end{tikzpicture}
& 
\begin{tikzpicture}
\begin{axis}[
    ybar,
    font=\scriptsize,
    title = {\roxf(H) and \rpar(H)},
    title style={yshift=-1ex},          
    width=0.5\linewidth,
    height=0.4\linewidth,
    legend style={opacity = 0.7, font=\tiny, row sep = -2pt, mark options={scale=3.5}, xshift = 0pt, cells={align=right}},
    legend cell align={right},
    legend pos=north east,
    grid=both,
    ylabel={Relative mAP improvement},
    xlabel={mAP per group before QE},
    symbolic x coords={A, B, C},
    xtick=data,
    xticklabels={$22.1$, $53.5$, $83.8$},
    % nodes near coords,
    % nodes near coords align={vertical},
    /pgf/bar width=3pt,% bar width
    xtick style={draw=none},
    ymin = 0, ymax = 128,
    ytick={0,13,...,130},
    x tick label style={inner sep=0pt},   
    enlarge x limits=0.3,
    ]
\addplot coordinates {(A, 84.95) (B, 3.01) (C, 3.28)};
\addplot coordinates {(A, 49.53) (B, 6.62) (C, 3.64)};
\addplot coordinates {(A, 50.99) (B, 6.69) (C, 2.40)};
\addplot coordinates {(A, 29.49) (B, 15.65) (C, 3.05)};
\addplot coordinates {(A, 119.09) (B, 13.44) (C, 5.29)};
\legend{AQE, AQEwD, DQE, $\alpha$QE, \ours}
\end{axis}
\end{tikzpicture}
\end{tabular}